# Opportunities and Challenges in Automatic Speech Recognition

Rashmi Makhijani
*Research Scholar,*
*GHRCE*
*Nagpur, India*

Urmila Shrawankar
*Asst. Prof., CSE Deptt.*
*GHRCE*
*Nagpur, India*

Dr. V. M. Thakare
*Prof. and HOD, CSE Deptt.*
*S. G. B. Amaravati University,*
*Amaravati, India*

**Abstract**

*Automatic speech recognition enables a wide range of current and emerging applications such as automatic transcription, multimedia content analysis, and natural human-computer interfaces. This paper provides a glimpse of the opportunities and challenges that parallelism provides for automatic speech recognition and related application research from the point of view of speech researchers. The increasing parallelism in computing platforms opens three major possibilities for speech recognition systems: improving recognition accuracy in non-ideal, everyday noisy environments; increasing recognition throughput in batch processing of speech data; and reducing recognition latency in realtime usage scenarios. This paper describes technical challenges, approaches taken, and possible directions for future research to guide the design of efficient parallel software and hardware infrastructures.*

## 1. Introduction

Applications in today's world can no longer rely on significant increases in processor clock rate for performance improvements, as clock rate is now limited by factors such as power dissipation [4]. Rather, parallel scalability (the ability for an application to efficiently utilize an increasing number of processing elements) is now required for software to obtain sustained performance improvements on successive generations of processors.

Automatic Speech Recognition (ASR) is an application that consistently exploits advances in computation capabilities. With the availability of a new generation of highly parallel single-chip computation platforms, ASR researchers are faced with the question of unlimited computing to make speech recognition better. The goal of the work reported here is to explore plausible approaches to improve ASR in three ways:

1. Improve Accuracy: Account for noisy and reverberant environments in which current systems perform poorly, thereby increasing the range of scenarios where speech technology can be an effective solution.

2. Improve Throughput: Allow batch processing of the speech recognition task to execute as efficiently as possible, thereby increasing the utility for multimedia search and retrieval.

3. Improve Latency: Allow speech-based applications, such as speech-to-speech translation, to achieve real-time performance, where speech recognition is just one component of the application.

This paper discusses current work as well as opportunities and challenges in these areas with regard to parallelization from the point of view of speech researchers.

## 2. Improving Accuracy

Speech recognition systems can be sufficiently accurate when trained with enough data having similar characteristics to the test conditions. However, there still remain many circumstances in which recognition accuracy is quite poor. These include moderately to seriously noisy or reverberant noise conditions, and any variability between training and recognition conditions with respect to channel and speaker characteristics (such as style, emotion, topic, accent, and language).

One approach that is both "embarrassingly" parallel and effective in improving ASR robustness is the socalled multistream approach. As has been shown for a number of years [5, 6, 15, 11], incorporating multiple feature sets consistently improves performance for both small and large ASR tasks. And as noted in [23], recent results have demonstrated that a larger number of feature representations can be particularly effective in the case of noisy speech. In order to conduct research on a massively parallel front end, a large feature space is desired. One approach that found to be useful is to compute spectro-temporal features. These features correspond to the output of filters that are tuned to certain rates of change in the time and frequency dimensions.

Various approaches have been devised to combine and select the inherently large number of potential spectro-temporal features because processing them entirely is currently considered computationally intractable.

### 2.1 Current Approach

Current preferred approach to robust feature extraction is to generate many feature streams with different spectro-temporal properties. For instance, some streams might be more sensitive to speech that varies at a slow syllabic rate (e.g., 2 per second) and others might be more sensitive to signals that vary at a higher rate (such as 6 syllables per second). The streams are processed by neural networks (Multi-Layer Perceptrons, or MLPs) trained for discrimination between phones and generate estimates of posterior phone probability distributions.

For MLP-based feature streams, the most common combining techniques are: (1) appending all features to a single stream; (2) combining posterior distributions by a product rule, with or without scaling; (3) combining posterior distributions by an additive rule, with or without scaling; and (4) combining posterior distributions by another MLP, which may also use other features.

Current best approach to combination is to train an additional Neural Network to generate combination weights by incorporating entropies from the streams as well as overall spectral information. A 28- stream system is used, including 16 streams from division of temporal modulation frequencies, 8 streams from division by spectral modulation frequencies, and 4 streams from a division by both [23]. Using this method, for the Numbers 95 corpus with the Aurora noises added [12] the average word error rate was 8.1%, reduced from 15.3% for MFCCs and first and second order time derivatives. While robustness to environmental acoustics is the main focus, four equally weighted streams, with quasi-tonotopically divided spectro-temporal features were used. The system yielded a 13.3% relative improvement on the baseline, lowering word error rate from 25.5% to 22.1%.

**2.2 Future Directions**

In the current approach, the same modulation filters are applied to the entire spectrum. Within this one feature stream, a pipe-and-filter parallel pattern can be used to distribute work across processing elements. Since the MLPs used within the stream depend on dense linear algebra, the wealth of methods to parallelize matrix operations can be exploited. The 28 streams can also be potentially expanded to hundreds or thousands of streams by applying the Gabor filters to different parts of the spectrum as separate streams using a map-reduce parallel pattern.

These techniques will be even more important to analyze speech from distant microphones at meetings, a task that naturally provides challenges due to noise and reverberation. Finally, there will be more parallelization considerations in combining the many stream methods with conventional approaches to noise robustness. As many stream feature combination naturally adapt to parallel computing architectures, the improvement will be significant.

### 3. Improving Throughput

Batch speech transcription can be "embarrassingly parallel" by distributing different speech utterances to different machines. However, there is significant value in improving compute efficiency, which is increasingly relevant in today's energy limited and form-factor limited devices and compute facilities.

The many components of an ASR system can be partitioned into a feature extractor and an inference engine. The speech feature extractor collects feature vectors from input audio waveforms using a sequence of signal processing steps in a data flow framework. Many levels of parallelism can be exploited within a step, as well as across steps, as described in section 2.1. Thus feature extraction is highly scalable with respect to the parallel platform advances. However, parallelizing the inference engine requires surmounting significant challenges.

The inference engine traverses a graph-based recognition network based on the Viterbi search algorithm [17] and infers the most likely word sequence based on the extracted speech features and the recognition network. In a typical recognition process, there are significant parallelization challenges in concurrently evaluating thousands of alternative interpretations of a speech utterance to find the most likely interpretation. The traversal is conducted over an irregular graph-based knowledge network and is controlled by a sequence of audio features known only at run time. Furthermore, the data working set changes dynamically during the traversal process and the algorithm requires frequent communication between concurrent tasks. These problem characteristics lead to unpredictable memory accesses and poor data locality and cause significant challenges in load balancing and efficient synchronization between processor cores. There have been many attempts to parallelize speech recognition on emerging platforms, leveraging both fine grained and coarse-grained concurrency in the application.

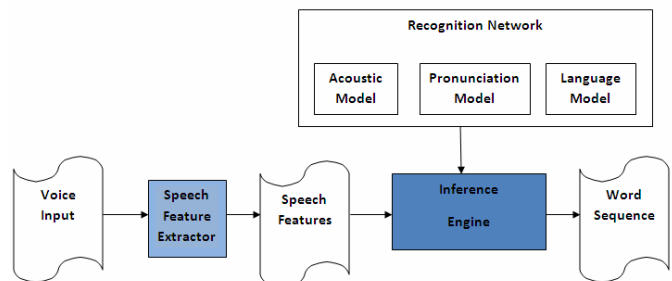

Fig. 1 Decoder Architecture

Fine-grained concurrency was mapped onto the multiprocessor with distributed memory in [20]. The implementation statically mapped a carefully partitioned recognition network onto the multiprocessors to minimize load imbalance. [14] explored coarse-grained concurrency in speech recognition and implemented a pipeline of tasks on a cellphone-oriented multicore architecture. [22] proposed a parallel speech recognizer implementation on a commodity multicore system using OpenMP. The Viterbi search was parallelized by statically partitioning a tree-lexical search network across cores. The parallel recognition system proposed in [19] also uses a weighted finite state transducer (WFST) and data parallelism when traversing the recognition network. Prior works such as [10, 7] leveraged many core processors and focused on speeding up the compute-intensive phase (i.e., observation probability computation) of ASR on many core accelerators. Both [10, 7] demonstrated approximately 5x speedups in the compute-intensive phase and mapped the communication intensive phases (i.e., Viterbi search) onto the host processor.

## 3.1 Current Approach

More recently, a data-parallel automatic speech recognition inference engine was implemented on the graphics processing unit (GPU), achieving over 11x speedup compared to SIMD optimized sequential implementation on an Intel core i7 CPU. With less than 8% sequential overhead, the solution promises more speedup on future more parallel platforms [8]. The speedup was enabled by constructing the recognition engine's software architecture to efficiently execute on single-chip manycore processors. There are four key implementation decisions that contributed to the speedup:

### 3.1.1 Exposing fine-grained parallelism

The software architecture of the inference engine is illustrated in Figure 1. The Hidden Markov model (HMM) based inference algorithm dictates that there is an outer iteration processing one input feature vector at a time. Within each iteration, there is a sequence of algorithmic steps implementing maximal-likelihood inference process. The parallelism of the application is inside each algorithmic steps, where the inference engine keeps track of thousands to tens of thousands of alternative interpretations of the input waveform. The challenge is that each algorithmic step only performs tens to hundreds of instructions on each alternative interpretation, thus synchronizations between the algorithmic steps impose sequential overheads. In multi-chip parallel platforms, the synchronization overhead significantly degrades parallel speedup. The opportunity brought by single-chip manycore parallel processors is that the synchronization overhead is significantly reduced to the point that the finegrained parallelism can be exposed and the application speedup potentials can be realized.

### 3.1.2 Implementing all parts of an algorithm on the GPU

Current GPUs are accelerator subsystems managed by a CPU over the PCIe data bus. As shown in Figure 1. In the inference engine, there is a compute intensive phase and a communication intensive phase of execution in each inference iteration. The compute intensive phase calculates the sum of differences of a feature vector against Gaussian mixtures in the acoustic model and can be readily parallelized. The communication intensive phase keeps track of thousands of alternative interpretations and manages their traversal through a complex finite state transducer representing the pronunciation and language models. While 17.7x speedup for the compute-intensive phase compared to sequential execution on the CPU was achieved , the communication-intensive phase is much more difficult to parallelize and received a 4.4x speedup. However, because the algorithm is completely implemented on the GPU, it has achieved a 11.3x speedup of the overall inference engine.

### 3.1.3 Leveraging fast hardware atomic operation support:

The inference process is composed of data-parallel graph traversals on the recognition network. The graph traversal routines are executing in parallel on difference cores and frequently have to update the same memory location. This causes race conditions as the same piece of data must be read and conditionally written by multiple instruction streams at the same time. The race condition can be resolved using a sequence of data parallel algorithmic steps in the application software or by using hardware-based atomic operation support. When leveraging hardware-based atomic operation support, however, the operations must be carefully managed as atomic operations to the same memory address are sequentialized.

### 3.1.4 Construct runtime data buffers to maximally regularize data access patterns:

The recognition network is an irregular network and the traversal through the network is guided by user input available only at runtime. In each iteration of the inference engine, to maximally utilize the memory load and store bandwidth, the data to be accessed is gathered during the iteration into a consecutive vector acting as runtime data buffers, such that the algorithmic steps in the iteration are able to load and store results one cache line at a time. This maximizes the utilization of the available data bandwidth to memory. With these four key implementation decisions, it is possible to overcome the parallelization challenges imposed by the application, and architect and implement a scalable parallel solution for speech recognition inference decoding.

## 3.2 Future Directions

The current work established an efficient software architecture for speech recognition targeting the highly parallel manycore platforms. The ongoing work is constructing an application framework that allows many additional features to be extended without jeopardizing the efficiency and throughput of the implementation. One example of such additional feature can be an alternative observation likelihood computation that reduces the amount of computation necessary. Other improvements to the software architecture include producing word lattices or confusion-networks in the context of multiple pass recognition systems. The improvements in recognition throughput could also be used to trade off speed with accuracy, making viable approaches.

## 4. Improving Latency

For speech recognition to be useful in multispeaker scenarios, it is also important to determine "who is speaking when", a process called "speaker diarization", and to further segment the speech in a way that is reasonable for human consumption.The ongoing research is by no means complete but speaker diarization is a good example for explaining the opportunities to improve latency.

Most speaker diarization systems use agglomerative hierarchical clustering as a core approach to perform diarization. At a high-level, systems extract MFCC features from a given audio track, discriminate between speech and nonspeech regions (speech activity detection), and use the agglomerative clustering approach to perform both segmentation of the audio track into speaker-homogeneous time segments and the grouping of these segments into speaker-homogeneous clusters in one step. Speech activity regions are determined using a speech/non-speech detector, e.g., [21]. The nonspeech regions are then excluded from the agglomerative clustering where the clustering is initialized using k clusters, with k larger than the number of speakers that are assumed to appear in the recording. Every cluster is modeled with a Gaussian Mixture Model containing g Gaussians. In order to train initial GMMs for the k speaker clusters an initial segmentation is generated by uniformly partitioning the audio into k segments of the same length. The ICSI system [1, 3] then performs the following iterations:

Re-Segmentation: Run Viterbi alignment to find the optimal path of frames and models. In the ICSI system, a minimum duration of 2.5 seconds is assumed for each speech segment. Re-Training: Given the new segmentation of the audio track, compute new Gaussian Mixture Models for each of the clusters. Cluster Merging: Given the new GMMs, try to find the two clusters that most likely represent the same speaker. This is done by computing a score based on the Bayesian Information Criterion (BIC) of each of the clusters and the BIC score of a new GMM trained on the merged segments for two clusters. If the BIC score of the merged GMM is larger than or equal to the sum of the individual BIC scores, the two models are merged and the algorithm continues at the re-segmentation step using the merged GMM. If no pair is found, the algorithm stops. As a result of different sequential optimization approaches [13], the current implementation runs at about 0.6× realtime, i.e., for 10 minutes of audio data, diarization finishes in roughly 6 minutes. The main problem with the approach is that it requires the complete recording of a meeting file and so the latency is the time of the meeting + 0.6× realtime of the meeting duration. There are many applications where online diarization is desirable and batch processing impractical.

### 4.1 Current Approach

An initial approach to online diarization was presented in the NIST Rich Transcription 2009 evaluations. The system consisted of a training step and an online recognition step. For the training step, the first 1000 seconds of the input are taken and performed offline speaker diarization using the system described above. Then speaker models are trained and a speech/non-speech model are taken from the output of the system. This is done by concatenating a random 60 second chunk of each speaker's segmented data and another one for the non-speech segments. In the online recognition step, the remainder of the meeting are recognized using the trained models. The sampled audio data is noise-reduced and converted into MFCC features. For every frame, the likelihood for each set of features is computed against each set of Gaussian Mixtures obtained in the training step, i.e. each speaker model and the non-speech model. A total of 250 ten ms frames is used for a majority vote on the likelihood values to determine the classification result. Therefore the latency totals at $t + 2.5$ s per decision (plus the portion of the offline training).

Such a system can significantly benefit from parallelism. If the offline diarization were two orders-of magnitude faster than realtime, the offline diarization could process one minute of meeting in less than a second.

### 4.2 Future Directions

Parallelism can be leveraged for low latency on different levels. The training of Gaussian Mixture Modes primarily requires matrix computation. If matrix computation is sped up by parallelism, more training can be run in the background at reduced wait times, resulting in both higher accuracy and lower latency. Also, giving models more iterations often leads them to converge with even less data, which also reduces latency. In the concrete example of diarization, lower runtime and therefore lower latency can be achieved by speeding up the cluster merge process, which might be parallelized on a thread level or using data parallelism by distributing each speaker model to a different core. With incoming data arriving through a sound card, USB device, or hard drive, I/O operations are likely to become a significant part of the runtime once parallelism is used intensively. Also, in the it was found that caching of highly repeated low-level operations (e.g., logarithm computations) helps runtime significantly. Therefore, a central cache for repeated operations seems highly desirable.

## 5. Conclusions

Automatic Speech Recognition (ASR) is an application that consistently benefits from more powerful computation platforms. With the increasing adoption of parallel multicore and manycore processors, significant opportunities for speech recognition can be seen in increasing recognition accuracy, increasing batch-recognition throughput, and reducing recognition latency. Here on-going work on these directions, focusing on the opportunities and challenges with regard to parallelization is described. The proposed directions for future research may serve to guide future designs of efficient parallel software and hardware infrastructures for speech recognition.